\documentclass[10pt,twocolumn,letterpaper]{article}

\usepackage{cvpr}
\usepackage{times}
\usepackage{epsfig}
\usepackage{graphicx}
\usepackage{amsmath}
\usepackage{amssymb}
\usepackage[ruled,vlined]{algorithm2e}

\usepackage[pagebackref=true,breaklinks=true,letterpaper=true,colorlinks,bookmarks=false]{hyperref}

\graphicspath{{.}{../figures/}}

\newcommand{\One}{\hbox{$1\hskip -1.2pt\vrule depth 0pt height 1.6ex width 0.7pt\vrule depth 0pt height 0.3pt width 0.12em$}}
\newcommand{\KKn}{\mathcal{K}_n}
\newcommand{\myargmax}[1]{\mathrm{arg}\hspace{-0.5mm}\max_{#1\phantom{ii}}}
\newcommand{\myyargmax}[1]{\mathrm{arg}\hspace{-4mm}\max_{#1\phantom{ii}}}

\newcommand{\comment}[1]{}

\cvprfinalcopy % *** Uncomment this line for the final submission

\def\cvprPaperID{1716} % *** Enter the CVPR Paper ID here

\ifcvprfinal\fi
\begin{document}

%%%%%%%%% TITLE
%\title{Autonomous Cleaning of Dirty Documents - A Generative Modelling Approach}
%
\title{Autonomous Cleaning of Corrupted Scanned Documents --\\ A Generative Modeling Approach}
%
%\title{Autonomous Cleaning of Corrupted Scanned Text Documents -\\ A Generative Modelling Approach}
%
%\title{A Probabilistic Generative Approach to the Cleaning of Corrupted Scanned Text Documents}
%
%\title{A Probabilistic Generative Approach to Dirty Document Cleaning}
%
\author{Zhenwen Dai\\
Frankfurt Institute for Advanced Studies,\\
Dept.\,of Physics, Goethe-University Frankfurt\\
{\tt\small dai@fias.uni-frankfurt.de}
\and
J\"{o}rg L\"{u}cke\\
Frankfurt Institute for Advanced Studies,\\
Dept.\,of Physics, Goethe-University Frankfurt\\
{\tt\small luecke@fias.uni-frankfurt.de}
}

%\IEEEoverridecommandlockouts
%\IEEEpubid{\makebox[\columnwidth]{\hfill 978-1-4244-8396-9/10/\$26.00~\copyright~2010~IEEE} \hspace{\columnsep}\makebox[\columnwidth]{ }}

\maketitle
% \thispagestyle{empty}

%%%%%%%%% ABSTRACT
\begin{abstract}
  We study the task of cleaning scanned text documents that are
  strongly corrupted by dirt such as manual line strokes, spilled ink
  etc.  We aim at autonomously removing dirt from a single letter-size
  page based only on the information the page contains. Our approach,
  therefore, has to learn character representations without
  supervision and requires a mechanism to distinguish learned
  representations from irregular patterns.
  To learn character representations, we use a probabilistic
  generative model parameterizing pattern features, feature variances,
  the features' planar arrangements, and pattern frequencies. The
  latent variables of the model describe pattern class, pattern
  position, and the presence or absence of individual pattern
  features. The model parameters are optimized using a novel
  variational EM approximation. After learning, the parameters
  represent, independently of their absolute position, planar feature
  arrangements and their variances. A quality measure defined based on
  the learned representation then allows for an autonomous
  discrimination between regular character patterns and the irregular
  patterns making up the dirt. The irregular patterns can thus be
  removed to clean the document.
  For a full Latin alphabet we found that a single page does not
  contain sufficiently many character examples. However, even if
  heavily corrupted by dirt, we show that a page containing a lower
  number of character types can efficiently and autonomously be
  cleaned solely based on the structural regularity of the characters
  it contains. In different examples using characters from different
  alphabets, we demonstrate generality of the approach and discuss its
  implications for future developments.
\end{abstract}

%%%%%%%%% BODY TEXT

\section{Introduction}
\ \vspace{-235mm}\\
%\phantom{wwwwwwwwwwwwwwwwwwwwwwiiiiiiiiii}
{\bf \makebox{978-1-4673-1228-8/12/\$31.00~\copyright~2012 IEEE}}\\[225mm]

A basic form of human communication, written text, consists of planar
arrangements of reoccurring and regular patterns. While in modern
forms of text these patterns are characters or symbols for
words (e.g., Chinese texts), early forms consisted of symbols
resembling objects.  Written text became a successful form of communication
because it exploits the readily available capability of the human
visual system to learn and recognize regular patterns in visual data.
In recent years, computer vision and machine learning became
increasingly successful in analyzing visual data. Much progress has
been made, for instance, by probabilistic modeling approaches that aim
at capturing the statistical regularities of a given data set.
Examples are image denoising by Markov Random Fields
\cite{Schmidt2010} or sparse coding models \cite{OlshausenField1996,LeeEtAl2007}.
%
%
% and many more), probabilistic
%approaches to subdivide images into meaningful segments (e.g., \cite{HeilerEtAl2005}), or
%the unsupervised learning of objects from visual data, e.g., by
%bilinear models (e.g., \cite{BerkesTurnerSahani2009,GrimesRao2005,TenenbaumEtAl2000}\comment{not sure?} or mixture-model approaches \cite{DudaHart1973}\comment{not sure?}).
%Many of these approaches are formulated
%as probabilistic generative models which aim at rebuilding the
%underlying data generation process.
%
For many types of data, modeling approaches hereby have to address the
problem that regular visual structures often appear at arbitrary
positions.  Sparse coding approaches indirectly address this problem by
replicating a learned structure (e.g., a Gabor wavelet) at different
positions of image patches. Other approaches go one step further and
explicitly model pattern positions using additional hidden variables
\cite{OlshausenEtAl1993,WiskottEtAl1997,Frey2003,GrimesRao2005,Williams2004a,Frey2005a}.
%Pattern representations can thus be learned independently of pattern
%positions.
However, the combinatorics of object identity and position
introduces major challenges as for each pattern class all positions
ideally have to be considered.

In this paper we apply a probabilistic generative approach with
explicit position encoding to remove dirt from text documents.  The
principle idea is very straight-forward: If characters are the salient
regular patterns of text, an appropriately structured probabilistic
model should be able to learn character representations as regular
arrangements of features. In contrast, dirt is much more irregular.
Coffee spots, spilled ink, or line-strokes scratching-out text share
similar features with printed characters but such corruptions are, on average,
much more random combinations of feature patterns.  Based on this
observation, the autonomous identification and recovery of characters
from a corrupted text document should thus be possible. But how
difficult is such a task? Or how robust can a solution of such a task
be if the data is heavily corrupted by dirt? Would the information
contained on a single page of a dirty document, for instance, be
sufficient to identify the characters containing it?  And if yes, can
this be used to `self-clean' the document? Such questions can, of
course, not be answered by a clear `yes' or `no' because they will,
e.g., depend on the type and degree of dirt or on the amount of
available character information on a page. However, we will show that
a self-cleaning of heavily corrupted documents is, indeed, possible,
e.g., for relatively low numbers of different character types. \comment{not sure about this sentence.$->$} The
only prerequisite will hereby be the characters' regular
feature arrangements. No information about the characters has to be available,
which makes the approach applicable to entirely unknown character
types. The problem addressed here is thus very different from the one
aimed at by optical character recognition (OCR) methods that use
supervised pretraining on known characters.
In contrast, we require unsupervised methods to learn character
representations. The generative model we apply is similar to models
suggested by Williams \& Titsias \cite{Williams2004a} and Jojic \& Frey
\cite{Frey2003,Frey2005a,Winn2004}, which provide explicit
representations of the data's regular patterns. As the data points we will
have to process are image patches of corrupted text documents, these
previous models are not applicable because they require a static
background, do not provide a mechanism to discriminate characters from
irregular patterns, and are based on pixel image representations which
can make learning less robust. In contrast, we (1)~will have to allow
for varying fore- and background patterns (to take dirt into account),
(2)~will introduce a mechanism for character vs.\ dirt discrimination,
and (3)~will consider general feature vector representations of the data.
Together with a novel non-greedy training scheme in the form of truncated
variational EM \cite{LuckeEggert2010}, the derived method will provide
the required robustness and efficiency for the task.
%
%
%\comment{The paper is structured as follows: We will first define the
%generative model (Sec.\,\ref{SecModel}) and derive an efficient
%algorithm for maximum likelihood estimation of the model parameters in
%Sec.\,\ref{SecOptimization}.  In Sec.\,\ref{SecLearning} we will show
%that the algorithm learns to represent regular patterns of artificial
%data and of scanned text documents.  Sec.\,\ref{SecClassification}
%describes how learned representations can be used for pattern
%identification and for the discrimination between character patterns
%and dirt.  Sec.\,\ref{SecExperiments} (and the Supplement) report
%experimental results.}
%
%
%Experimental results on the of the complete procedure applied to tasks of document cleaning
%are finally reported in Sec.\,\ref{SecExperiments}.
%
%
%
%
%\vspace{-0.3in}
%
%
%
\section{A Generative Model for Characters}
\label{SecModel}
The probabilistic model we consider generates small image patches of
size $ \vec{D} = (D_1,D_2)$.  A pixel at position $\vec{d}$ of
the patch is represented by a feature vector $\vec{y}_{\vec{d}}$ with
$F$ entries.  For now $\vec{y}_{\vec{d}}$ can be thought of as a color
vector at pixel position $\vec{d}$ in RGB space ($F=3$). For the
application to text documents we will later use more sophisticated
features, however.

\begin{figure}[t]
\begin{minipage}[b]{1.0\linewidth}
\begin{center}
\includegraphics[width=1.0\linewidth]{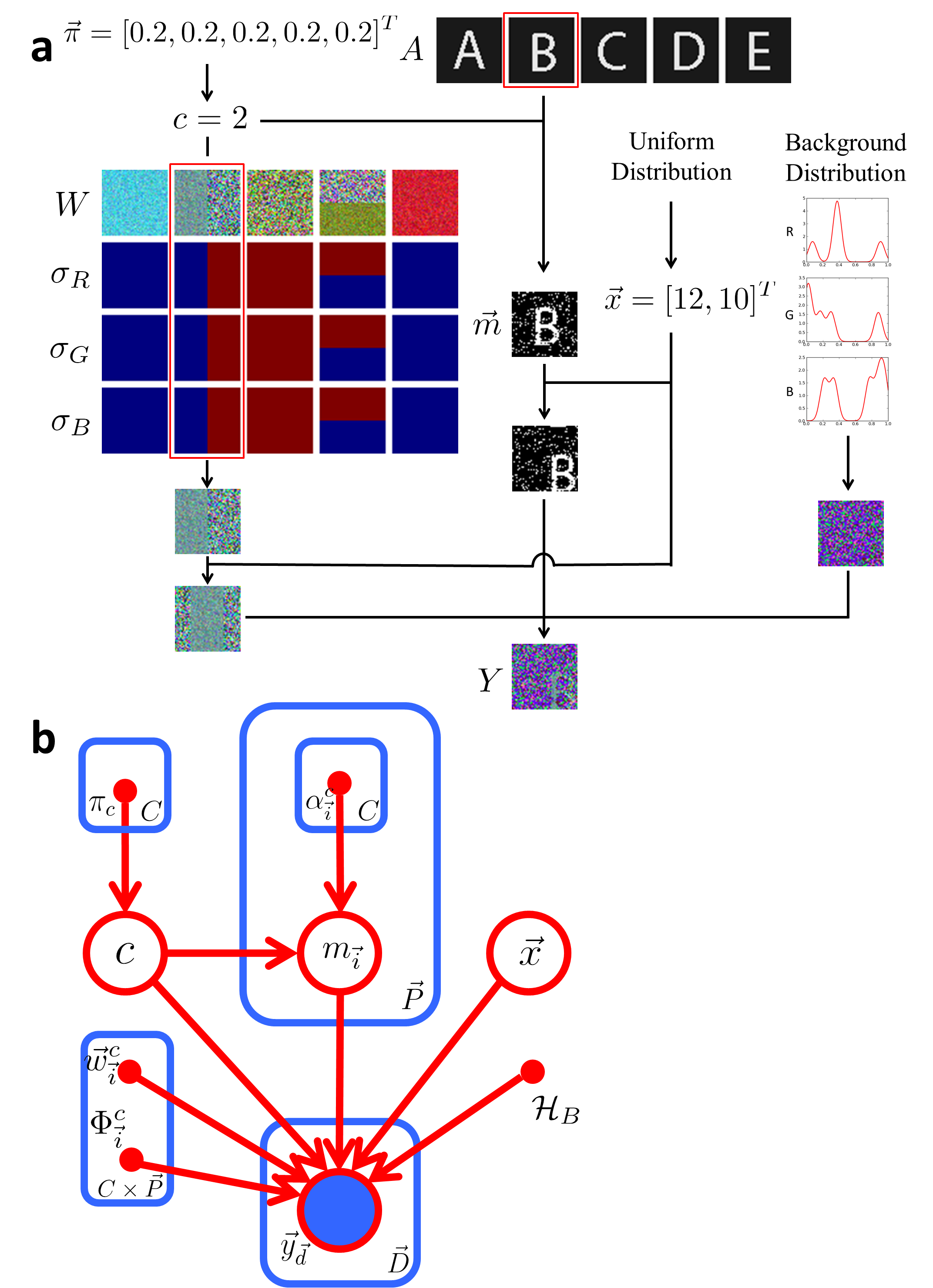}
\end{center}
\end{minipage}
\vspace{-2.0cm}
\begin{flushright}
\begin{minipage}[b]{0.35\linewidth}
\caption{\textbf{a} Illustration of the generation process. \textbf{b} The graphical model.}\label{fig:model}
\end{minipage}
\end{flushright}\vspace{-3mm}
\end{figure}

A patch $Y=(\vec{y}_{(1,1)},\ldots,\vec{y}_{(D_1,D_2)})$ is modeled to
contain one pattern at an arbitrary position of the patch (see
Fig.\,\ref{fig:model}a). For the class variable $c$ we use a
standard mixture model prior with $\vec{\pi}=(\pi_1, \ldots, \pi_C)$
denoting the mixing proportions and $C$ denoting the total number of classes:
{\small
\begin{equation}
p(c|\vec{\pi}) = \textstyle\pi_c \qquad \text{with} \quad \sum_{c=1}^C \pi_c = 1\,. \label{eqn_piprior}
\end{equation}}
The position of the pattern in the patch, $\vec{x}\in{\cal D}, \mathcal{D} = \{1,\ldots,D_1\}\times\{1,\ldots,D_2\}$, is a 2D vector chosen from a uniform distribution over the entire patch: \vspace{-2mm}
{\small
\begin{equation}
\begin{split}
p(\vec{x}) &= p(x_1) p(x_2)\\ &= \text{Uniform}(1,D_1)\,\times\,\text{Uniform}(1,D_2) = \frac{1}{D_1D_2}.
\end{split}
\end{equation}
}
The shapes of different patterns are modeled by a set of binary latent variables, namely the pattern {\em mask}:
$\vec{m}= ( m_{(1,1)}, \ldots, m_{(P_1,P_2)})$, where $m_{\vec{i}} \in \{0, 1\}$. With $m_{\vec{i}}=1$
the corresponding feature is part of the pattern, while with $m_{\vec{i}}=0$ it is part of the background. The pattern size $\vec{P} = (P_1,P_2)$ can be different from the image patch size $P_1 \leq D_1, P_2 \leq D_2$. Given the pattern class $c$, the mask variables are drawn from Bernoulli distributions:
{\small
\begin{equation}
p(\vec{m} | c, A) = \prod_{\vec{i}=(1,1)}^{(P_1,P_2)} p(m_{\vec{i}}| c, A)
%= \prod_{\vec{i}=(1,1)}^{(P_1,P_2)} \text{Bernoulli}(m_{\vec{i}} ; \alpha_{\vec{i}}^c)
= \prod_{\vec{i}=(1,1)}^{(P_1,P_2)} (\alpha_{\vec{i}}^c)^{m_{\vec{i}}} (1-\alpha_{\vec{i}}^c)^{1-m_{\vec{i}}}, \label{eqn_maskprior}
\end{equation}
}
where $A = (A^1, \ldots, A^C)$ with $A^c = (\alpha^c_{(1,1)}, \ldots, \alpha^c_{(P_1,P_2)})$ are the parameters of the mask distribution. For the area where the image patch is outside the pattern, the mask variables
are always assigned $0$: $p(m_{\vec{d}}=1\,|\,c,A) = p(m_{\vec{d}}=1) = 0, \forall \vec{d} \in \mathcal{D} -\mathcal{P}, \mathcal{P} = \{1,\ldots,P_1\}\times\{1,\ldots,P_2\}$.
%
%\begin{equation}
%p(m_{\vec{d}}) = 0, \forall \vec{d} \in \mathcal{D} -\mathcal{P}, \mathcal{P} = \{1,\ldots,P_1\}\times\{1,\ldo%%ts,P_2\}. \label{eqn_maskextra}
%\end{equation}
%
From the definition of masks, a background distribution is required for all those features not belonging to a pattern ($m_{\vec{i}}=0$).
%It serves the purpose of modeling the distribution of image features where no regular patterns exist.
A possible choice is a flat Gaussian distribution (compare \cite{Williams2004a}).
%
%, which is suitable for the evenly distributed values, e.g., grey-scale image intensities.
%
However, for data such as patches from corrupted text documents, the distribution values are often
very different for the different feature vector entries, and for the dirty background are often
observed to be non-Gaussian.
%
%show very multimodal
%However, the distributions of feature values vary very much among different image features, especially those edge-based features e.g. Gabor, which is similar to heavy-tailed distributions.
%
To appropriately model the background features, we therefore construct
a probability density function $\mathcal{H}_B$ by computing the
histogram of different feature values across the image
patches.  The
probability densities for the individual feature vector entries will
be modeled individually (see Fig.\,\ref{fig:artificialdata}a for histograms of
R, G, and B channel). The histograms are computed across all the image
patches including the features that are potentially later
identified as being part of the learned patterns.  Nevertheless, the
computed histograms are usually very similar to the true background
distributions (compare Fig.\,\ref{fig:artificialdata}a). Once computed we therefore leave
the histograms fixed throughout learning.
%
%In principle, the background histograms could also be updated by deriving rules similar to the above ones. However, since the background model describes the general distribution of features, it should not be significantly changed during the learning procedure. In our method, for simplicity, the background histogram is fixed to the initial statistics over the whole data set.
%
Having defined the background distribution $\mathcal{H}_B$ and given
pattern class $c$, mask $\vec{m}$, and pattern position $\vec{x}$, the
distribution of patch features is given by:
{\small
\begin{equation}
\begin{split}
p( Y | c, \vec{m}, \vec{x}, \Theta) = &\prod_{\vec{d}=(1,1)}^{(D_1,D_2)} \left[ m_{(\vec{d}-\vec{x})} \mathcal{N}(\vec{y}_{\vec{d}}\;; \vec{w}_{(\vec{d}-\vec{x})}^c, \Phi^c_{(\vec{d}-\vec{x})})\right. \\ & \left.+(1-m_{(\vec{d}-\vec{x})}) \mathcal{H}_B(\vec{y}_{\vec{d}}) \right], \label{eqn_likelihood}
\end{split}\hspace{-3mm}
\end{equation}
}
where $\vec{w}^c_{\vec{i}}$ is the mean of a Gaussian distribution and $\Phi^c_{\vec{i}}$ is the
diagonal convariance matrix: $\Phi^c_{\vec{i}} = \mathrm{diag}\big((\sigma^c_{\vec{i}, f=1})^2,\ldots,(\sigma^c_{\vec{i}, f=F})^2\big)$.
%\[
%\Phi^c_{\vec{i}} =
%\begin{pmatrix}
%  (\sigma^c_{\vec{i}, f=1})^2 &  & & \\
%  & (\sigma^c_{\vec{i}, f=2})^2 & & \\
%  & & \ddots & \\
%  & & & (\sigma^c_{\vec{i}, f=F})^2
%\end{pmatrix}.
%\]
The mean $\vec{w}^c_{\vec{i}}$ parameterizes the mean feature vector of pattern $c$
at position $\vec{i}$ relative to the pattern position $\vec{x}$. The variance vector
$\Phi^c_{\vec{i}}$ parameterizes the feature vector variances (different variance per
vector entry). The shift of a pattern $c$ is implemented by a change of the position
indices $\vec{i}$ by $\vec{x}$ using cyclic boundary positions:
{\small
\begin{equation}\label{eqn_cyclic_boundary}
\vec{d} = (\vec{i}+\vec{x}):=\Big( (i_1+x_1) \bmod D_1, (i_2+x_2) \bmod D_2 \Big)^T.
\end{equation}
}
Equations \ref{eqn_piprior} to \ref{eqn_cyclic_boundary} define the
generative model for image patches.  The parameters of the model are
given by $\Theta = (W, \Phi, A, \vec{\pi})$ with $W=(W^0, \ldots, W^C)$ and
$W^c=(\vec{w}^c_{(1,1)}, \ldots, \vec{w}^c_{(P_1,P_2)})$, and together with the histograms for
the background distribution. Fig.\,\ref{fig:model}a shows schematically
how a patch is generated.  First, a pattern class is chosen (e.g.,
the class with pattern ``B") and then the mask variables $\vec{m}$ for the class
(Eqn.\,\ref{eqn_maskprior}). Pattern parameters and mask are then
translated by a random position $\vec{x}$ before they are combined
through a Gaussian distribution for the model and the learned
distribution for the background Eqn.\,\ref{eqn_likelihood}.

% a position $\vec{x}$ and
%
%
%
%Learning the shape of the object could be very beneficial. It could th We do not want the parameters outside the true object area disturbing the calculation of the likelihood probability, since they are actually not part of the object.
% [discussion] Frey and Jojic paper does not explicitly use the mask. In their exp, those area will be averaged out, since they are not constant under translations, which usually results with flat means and a large variance. However, if the image feature go beyond image intensities, it will not be the case, because a Gaussian is a proper background model for complicated image features, as shown in the background probability distribution discussion.
% About mask, it could not only define the object shape, but also adapts to image when the object is not complete.

% [TODO] write about the selection of object in mixture model.

% [discussion] the several choices of background probabilities in Frey Jojic and williams papers
% [discussion] the general usage of the background distribution

%It is a mixture model.
%a set of objects
%the background
%the translation variable x
% - relation with control unit (introduction)
% - circular space
% - relation with permutation matrix and extension to general transformation

\section{Efficient Likelihood Maximization}
\label{SecOptimization}
One approach of learning the parameters $\Theta$ from data $\mathcal{Y} = (Y^{(1)},\ldots,Y^{(N)})$ is to maximize the data likelihood:
{\small
\begin{equation}
\begin{split}
\Theta^{*} &= \arg\max_{\Theta} \{\mathcal{L}(\Theta)\},\\ \quad \mathcal{L}(\Theta)&=\log\big(p(Y^{(1)},\ldots,Y^{(N)}|\Theta)\big).
\end{split}
\end{equation}
}
A frequently used method to find the parameters $\Theta^{*}$ is
Expectation Maximization (EM), which iteratively optimizes a lower
bound of the likelihood $\mathcal{F}(\Theta,q)$ w.r.t.\ the parameters $\Theta$ and a distribution~$q$. With $\sum_V$ denoting a summation across the joint space of all hidden variables $V=(c,\vec{m},\vec{x})$
it is given by:
{\vspace{-1mm}\small
\begin{align} \label{EqnFreeEnergy}
\mathcal{F}(\Theta,q) = &\hspace{-0.5mm}\sum^N_{n=1} \sum_V q_n(V, \Theta') \left[ \log(p(Y^{(n)}|V,\Theta))+\log(p(V|\Theta)) \right] \nonumber\\
&-\sum_{n=1}^N \sum_V q_n(V,\Theta') \log(q_n(V,\Theta')).%\hspace{-0.5mm}\vspace{-20mm}
\end{align}\vspace{-1mm}
}
%{\footnotesize
%\begin{equation}
%\label{EqnFreeEnergy}
%\begin{split}
%\mathcal{F}(\Theta,q) = &\hspace{-0.5mm}\sum^N_{n=1} \sum_V q_n(V, \Theta') \left[ \log(p(Y^{(n)}|V,\Theta)) \right.\\
%&\left.+\log(p(V|\Theta)) \right] -\sum_{n=1}^N \sum_V q_n(V,\Theta') \log(q_n(V,\Theta')).\hspace{-0.5mm}
%\end{split}\hspace{-3mm}
%\end{equation}
%}
%\begin{equation}
%H(q)=-\sum_{n=1}^N \sum_V q_n(V,\Theta') \log(q_n(V,\Theta')).
%\end{equation}
%
\textbf{M-Step.} Parameter update rules are canonically derived by setting the derivatives of $\mathcal{F}$ w.r.t. the parameters to $0$. For the model (\ref{eqn_piprior}) - (\ref{eqn_likelihood}), we obtain:
\begin{equation}
\hspace{-0.1in}
\begin{array}{l @{\;} l}
\pi_c = \frac{1}{N}\sum\limits_n \sum\limits_{\vec{x}} p^{(n)}_{\Theta}(c,\vec{x}), \\
\alpha^c_{\vec{i}} = \frac{\sum\limits_n \sum\limits_{\vec{x}} p^{(n)}_{\Theta}(c,\vec{x}) p^{(n)}_{\Theta}(m_{\vec{i}}=1|c,\vec{x})}{\sum\limits_n \sum\limits_{\vec{x}} p^{(n)}_{\Theta}(c,\vec{x})}, \\
\vec{w}^c_{\vec{i}} = \frac{\sum\limits_n \sum\limits_{\vec{x}} p^{(n)}_{\Theta}(c,\vec{x}) p^{(n)}_{\Theta}(m_{\vec{i}}=1|c,\vec{x}) \vec{y}^{(n)}_{(\vec{i}+\vec{x})}}{\sum\limits_n \sum\limits_{\vec{x}} p^{(n)}_{\Theta}(c,\vec{x}) p^{(n)}_{\Theta}(m_{\vec{i}}=1|c,\vec{x})}, \\
\Phi^c_{\vec{i}} = \frac{\sum\limits_n \sum\limits_{\vec{x}} p^{(n)}_{\Theta}(c,\vec{x}) p^{(n)}_{\Theta}(m_{\vec{i}}=1|c,\vec{x}) \big( (\vec{y}^{(n)}_{(\vec{i}+\vec{x})}-\vec{w}^c_{\vec{i}}) (\vec{y}^{(n)}_{(\vec{i}+\vec{x})}-\vec{w}^c_{\vec{i}})^T \odot\,\One \big)}{\sum\limits_n \sum\limits_{\vec{x}} p^{(n)}_{\Theta}(c,\vec{x}) p^{(n)}_{\Theta}(m_{\vec{i}}=1|c,\vec{x})}, \\
\end{array} \label{eqn_mstep}
\hspace{-0.5in}
\end{equation}
%where we used the abbreviations
where we abbreviated: $p^{(n)}_{\Theta}(m_{\vec{i}}|c,\vec{x}) := p(m_{\vec{i}}|Y^{(n)},c,\vec{x},\Theta),\ p^{(n)}_{\Theta}(c,\vec{x}) := p(c,\vec{x}|Y^{(n)},\Theta)$, and where $\odot$ denotes pointwise matrix multiplication (in this case with the unit matrix).
%
% speed issue: tractable but very slow

% [MAYBE] explain what is pixel independence assumption
\noindent\textbf{E-Step.} The crucial and computationally expensive part of EM
is the computation of the expectation values w.r.t.\ the posterior.
For each data point, this involves summations of probabilities for all
combinations of the hidden variables $c$, $\vec{m}$ and $\vec{x}$.
However, the summation over the latent combinations can be
simplified.  By exploiting the standard assumption of independent
observed variables (compare, e.g.,
\cite{OlshausenField1996,LeeEtAl2007}) given the latents
(see Eqn.\,\ref{eqn_likelihood}), the posterior distribution over $\vec{m}$ can be decomposed into the product of the posteriors over individual binary masks as follows:
{\small
\begin{equation}
\hspace{-3mm}p(c,\vec{m},\vec{x}|Y,\Theta) = \Big( \prod_{\vec{i}=(1,1)}^{(P_1,P_2)} p(m_{\vec{i}}|Y,c,\vec{x},\Theta) \Big) p(c,\vec{x}|Y,\Theta). \label{eqn_posteriori}
\end{equation}
}
The posterior distribution over individual binary masks can then be computed as follows:
{\small
\begin{equation}
p(m_{\vec{i}}|Y,c,\vec{x},\Theta) = \frac{p(\vec{y}_{(\vec{i}+\vec{x})},m_{\vec{i}}|c,\vec{x},\Theta)}{\sum_{m_{\vec{i}}'} p(\vec{y}_{(\vec{i}+\vec{x})},m_{\vec{i}}'|c,\vec{x},\Theta)}. \label{eqn_q2}
\end{equation}
}
The summation in the denominator can be computed efficiently as it only contains two cases: $m_{\vec{i}}=0$ and $m_{\vec{i}}=1$. The posterior distribution over $c$ and $\vec{x}$ can be computed as follows,
{\small
\begin{equation}
\begin{split}
p(c,\vec{x}|Y,\Theta) \propto &\big[ \prod_{\vec{i}=(1,1)}^{(P_1,P_2)} \big( \sum_{m_{\vec{i}}} p(\vec{y}_{(\vec{i}+\vec{x})},m_{\vec{i}}|c,\vec{x},\Theta) \big) \big]\\ &\cdot p(\vec{x}|\Theta)p(c|\Theta). \label{eqn_q_xc}
\end{split}
\end{equation}
}
With such a decomposition (compare \cite{Frey2005a,Williams2004a}), the computational complexity decreases from exponential to polynomial, which makes the computation tractable in principle. However, the computational complexity still grows very fast with the size of patterns and patches, $\mathcal{O}(C D_1 D_2 P_1 P_2)$. For realistic image sizes (e.g., usually hundreds of thousands of pixels), it still exceeds currently available computational resources.
%
%
%too slow for most vision applications such as object recognition.
%
%\textbf{Truncated variational EM.}
% [discussion] the comparison with the FFT-speedup approach
%
To further improve efficiency, we therefore approximate the computation of
expectation values using variational EM (e.g., \cite{JordanEtAl1999}). Source of the large computation is
the required evaluation of all possible pattern positions for all
classes. %The state space is large although relatively low dimensional.
To reduce the number of hidden states that have to be evaluated, we
apply a recent variational EM approach (Expectation Truncation, \cite{LuckeEggert2010}) which is well
suited for discrete hidden variables. The used approach is not based on the usual factored form of $q$
but on a truncated variational approximation to the
posterior. Applied to the posterior (\ref{eqn_q_xc}) it is given by:\vspace{-2mm}
{\small
\begin{equation}
\label{EqnETPosterior}
\begin{split}
p(c,\vec{x}\,|\,Y^{(n)},\Theta)&\approx{}q_n(c,\vec{x};\Theta)\\&=\frac{p(c,\vec{x},Y^{(n)}\,|\,\Theta)}{\sum_{(c,\vec{x})\in\KKn}p(c,\vec{x},Y^{(n)}\,|\,\Theta)}, \forall (c,\vec{x})\in\KKn,
\end{split}
\end{equation}
}
and zero otherwise. The variational distribution $q_n$ approximates
the true posterior with high precision if the set $\mathcal{K}_n$
contains those classes and positions that carry most posterior mass
for a given data point $Y^{(n)}$.  In other words, for a given patch
we have to find the most likely pattern classes together with their
most likely patch positions in order to obtain a high quality
approximation. To achieve this we define a function
$\mathcal{S}_{\Theta}^{(n)}(c,\vec{x})$ that assigns a score to each
class and position pair $(c,\vec{x})$:
{\small
\begin{equation}
\begin{split}
\mathcal{S}_{\Theta}^{(n)}(c,\vec{x}) =&\hspace{-2mm} \prod_{\vec{i}' \in \mathcal{P}'_c} \Big[ \mathcal{N}(\vec{y}_{(\vec{i}'+\vec{x})}^{(n)}\;; \vec{w}_{\vec{i}'}^c, \Phi^c_{\vec{i}'})p(m_{\vec{i}'}=1|\Theta)\\ + &\mathcal{H}_B(\vec{y}_{(\vec{i}'+\vec{x})}^{(n)}) p(m_{\vec{i}'}=0|\Theta) \Big] p(\vec{x}|\Theta) p(c|\Theta),\label{EqnSelection}
\end{split}
\end{equation}
}
with $\mathcal{P}'_c \subseteq \mathcal{P}$. This scoring (or selection) function (compare
\cite{LuckeEggert2010}) gives high values to all those positions that
are consistent with features in the set $\mathcal{P}'_c$.  The set
$\mathcal{P}'_c$ is in turn defined to contain the $\lambda$ most
reliable features of pattern $c$. We define these features as those
with the highest mask parameters $\alpha^c_{\vec{i}}$. A small number
of $\lambda$ results in a very efficiently computable function $\mathcal{S}_{\Theta}^{(n)}(c,\vec{x})$.
%
%For low
%numbers of
%
%
%Eqn.\,\ref{EqnSelection}
%we thus still have to evaluate all combination $(c,\vec{x})$ but
%each combination can be
%
%
Based on the selection function, we now define the set of most likely class and position pairs to be:
{\small
\begin{equation}
\begin{split}
\mathcal{K}_n = &\{(c,\vec{x}) | (c,\vec{x}) \text{ has one of}\\ &\text{the $(K\,C\,D_1\,D_2)$ largest values of } \mathcal{S}_{c,\Theta}^{(n)}(\vec{x}) \},
\label{EqnKKn}
\end{split}
\end{equation}
}
where $K\in[0,1]$ is the fraction of the joint space of all classes and positions (size $C\,D_1\,D_2$).

In principle, the approximation \cite{LuckeEggert2010} can also be
used to constrain the number of states of mask variables. However,
the computational gain is negligible as the posterior w.r.t.\ the mask can
be computed efficiently (\ref{eqn_q2}). For the approximation, note
that $\lambda$ and $K$ parameterize the accuracy. The higher $\lambda$ the more reliably is the selection of
considered classes and positions, and the higher $K$ the larger is the considered area of
the joint class and position space. However, the larger $\lambda$ and $K$ the
higher is the computational cost. For the highest possible value of
$\lambda$ the selection becomes optimal as $\mathcal{S}_{\Theta}^{(n)}(c,\vec{x})$
becomes proportional to $p(c,\vec{x}\,|\,Y^{(n)},\Theta)$ ($\mathcal{S}_{\Theta}^{(n)}(c,\vec{x})$ becomes equal to $p(c,\vec{x}\,|\,Y^{(n)},\Theta) p(Y^{(n)}\,|\,\Theta)$ with $p(Y^{(n)}\,|\,\Theta)$ being a constant for the selection).  For
the highest possible value of $K$, $K=1$, all positions are considered
and the variational distribution (\ref{EqnETPosterior}) becomes equal
to the exact posterior. In numerical experiments we found
approximations with high accuracy and simultaneously low computational
costs by choosing relatively low numbers of $\lambda$ (e.g., $\lambda=200$ out of $P_1P_2$ features)
and relatively low fractions of considered joint space (e.g., $K=0.02$).
%
% [TODO] the randomness in selecting D'_c

% [TODO] initialize As
%\begin{algorithm}
%\caption{ the algorithm flow with ET}
%\DontPrintSemicolon
%Choose approximation parameters $K$ and $\gamma$ and randomly initialize the parameters of the generative model. \;
%\While{parameters have not converged}{
%	Select the representative feature index set $D'_c$ for each object $c$. \;
%	\For{ all data points n=1,\ldots,N} {
%	Compute $\mathcal{K}_n$ by evaluating $\mathcal{S}_{\Theta}^{(n)}(\vec{x},c)$ on the joint space of $\vec{x}$ and $c$.\;
%	Compute the posterior distribution (\ref{eqn_posteriori}) over the truncated set $\mathcal{K}_n$.\;
%	}
%	Update the parameters in the M-step (\ref{eqn_mstep}) using the approximate posterior distribution.
%}
%\end{algorithm}

\section{Learning and Identification of Characters}\label{SecLearning}
Equations (\ref{eqn_mstep}) to (\ref{EqnKKn}) define an approximate
EM algorithm to learn character representations. These representations
will be used to remove dirt from documents as described in this section.
Before, we numerically evaluate the learning procedure itself.\\[1mm]
\begin{figure*}[t]
\begin{minipage}[b]{12.6cm}
\begin{center}
 \includegraphics[width=12.5cm]{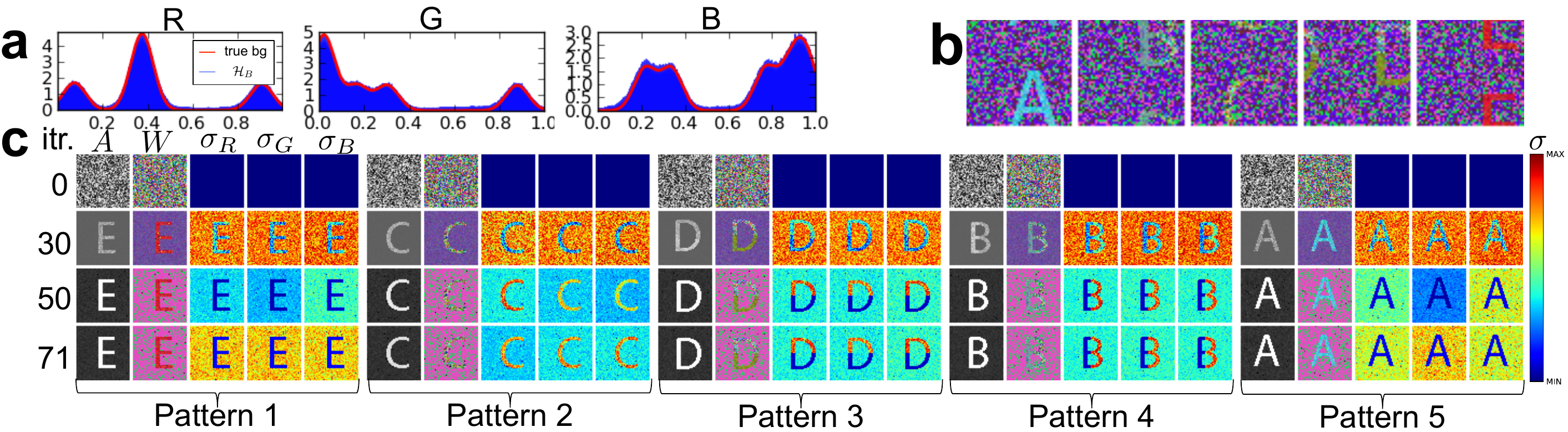}
\end{center}
\end{minipage}
\raisebox{.2cm}{
\begin{minipage}[b]{4.5cm}
\caption{Experiment on artificial data. \textbf{a} The background for data generation (red curves) and constructed histogram $\mathcal{H}_B$ (blue regions). \textbf{b} 5 of $N=1000$ image patches. \textbf{c} The learning course of the parameters. $W$ is visualized in RGB color
space and $\sigma_R$, $\sigma_G$ and $\sigma_B$ are visualized by heat maps.}\vspace{-2mm}
\label{fig:artificialdata}
\end{minipage}
}\vspace{-3mm}
\end{figure*}
% experiment setup
\noindent{}{\bf Artificial data.} Let us first consider artificial images for
which ground truth information is available.
For the training data, we generated $N=1000$ RGB image patches ($F=3$) of size $\vec{D}=(50,50)$ according to the model (\ref{eqn_piprior}) to (\ref{eqn_likelihood}). Each patch contained one of five different character types with equal probability ($\pi_c=0.2$).
The chosen colored characters were generated from corresponding mask, mean and variance parameters (see Fig.\,\ref{fig:model}a).
The background color was drawn from a Mixture of Gaussians as an example of multi-modal distributions (see Fig.\,\ref{fig:model}a). Fig.\,\ref{fig:artificialdata}b shows a random selection of $5$ generated data points.
% algorithm initialization
The derived EM learning algorithm was applied to the data assuming $C=5$ classes and $\vec{P}=\vec{D}=(50,50)$. First, the background histogram $\mathcal{H}_B$ was computed from the whole data set, and was observed to model the
true generating distributions with high accuracy (the blue regions in Fig.\,\ref{fig:artificialdata}a
show the learned histograms compared to the true distributions in red). To infer the remaining model parameters
they were first initialized: the pattern mean $W$ was independently and uniformly drawn from the RGB-color-cube $[0,1]^3$; the pattern variance $\Phi$ was set to the standard deviation of the data set; and the initial mask parameters $A$ were uniformly drawn from the interval $[0,1]$.
% comments on results
The learning course of the parameters is illustrated in Fig.\,\ref{fig:artificialdata}c with iteration $0$ showing the initial values. After iteration $70$, parameters had converged sufficiently.
To visualize pattern variances in Fig. \ref{fig:artificialdata}c, they are organized as a matrix for each pattern and each feature dimension, e.g.\ $\sigma_R^c = (\sigma^c_{(1,1),f=R}, \ldots, \sigma^c_{(P_1,P_2),f=R})$. These variance matrices are visualized by color images which are normalized individually. As can be observed,
the algorithm successfully learned the model parameters. For the experiment of Fig.\,\ref{fig:artificialdata}
and other similar experiments, the learned parameters diverged from the generating parameters by on average
less than $3.0\%$. Convergence to local optima has only been observed in very few cases ($1$ of $10$ runs).

\begin{figure*}[t]
\begin{center}
 \includegraphics[width=16cm]{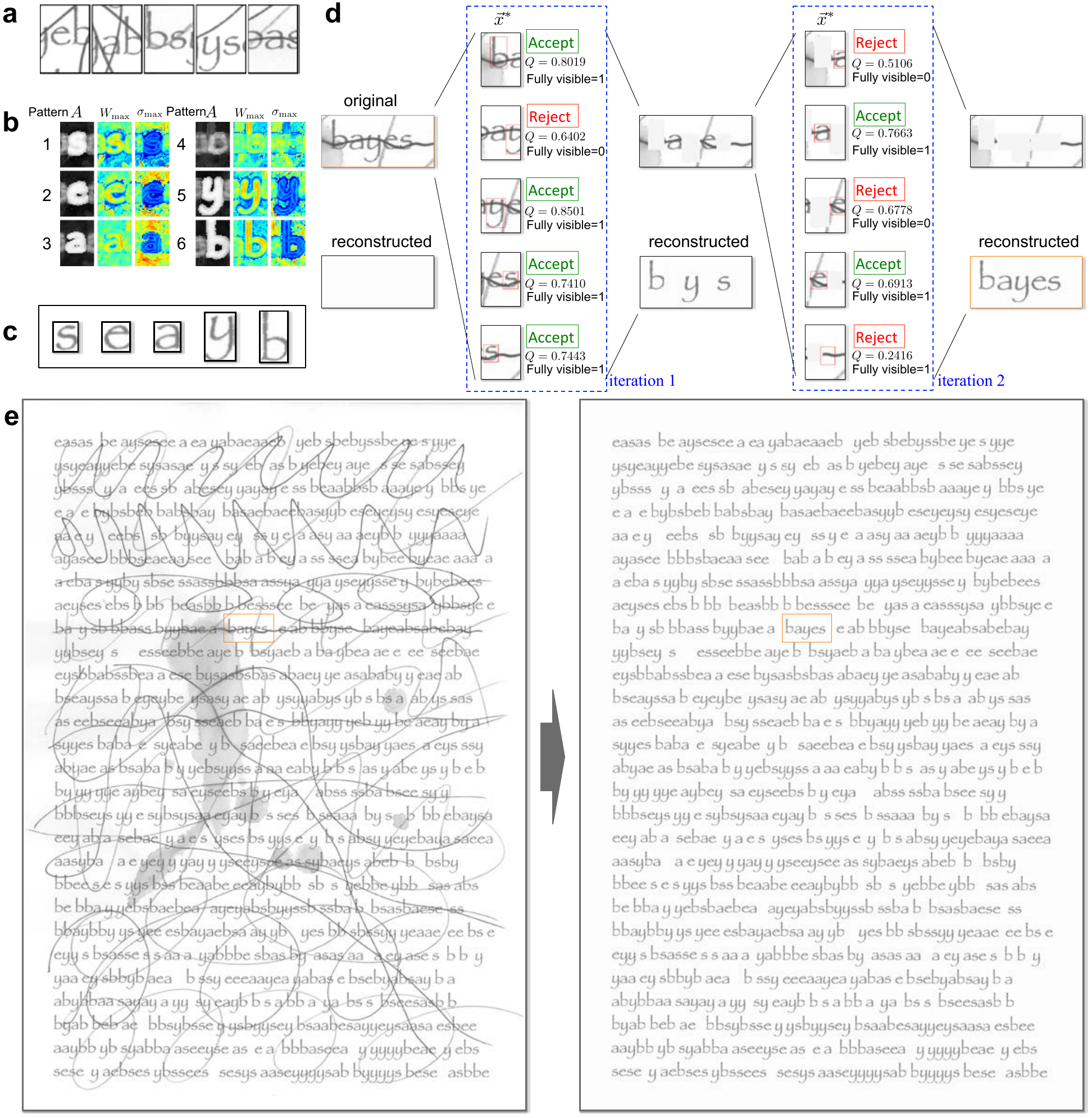}
\end{center}
\caption{Experiment on text document. \textbf{a} $5$ of $N=1379$ image patches. \textbf{b} The learned parameters with max representation,e.g., $W^c_{\vec{i}, \max}=\max_{f}(W^c_{\vec{i}, f})$ (see Suppl. for the full representation). \textbf{c} The clean representation of each character type. \textbf{d} An illustration of the cleaning procedure. \textbf{e} The cleaning result of our algorithm.}
\label{fig:bayes}
\vspace{-3mm}
\end{figure*}

\normalsize
\noindent{}{\bf Scanned text documents.} Let us now apply the learning algorithm to data from a single page
of a scanned text document. Consider the corrupted document displayed in Fig.\,\ref{fig:bayes}e
which contains $5$ character types, ``a", ``b", ``e", ``s" and ``y".
The printed document was manually corrupted with dirt in the form of line-strokes and
with grayish spots. The dataset for training was created by a high-resolution scan of the document ($3307\times{}4677$ pixels) and by automatically cutting the scan into small patches ($120\times165$ pixels) with fixed intervals. Fig.\,\ref{fig:bayes}a shows five examples of such patches. The patches are used to generate the
actual data points $Y^{(n)}$ with vectorial features. Instead of RGB feature vectors as for the introductory
example, we used feature
vectors generated through Gabor filter responses\comment{(see Supplement)}. Gabor features are robust and widespread in image processing (see, e.g., \cite{WiskottEtAl1997,Shen2006}) with high sensitivity to edge-like structures and textures. Furthermore, they are tolerant w.r.t.\ small local deformations and brightness changes.
For the small patches we computed a Gabor feature with $40$ entries at every third pixel,
which resulted in 2D arrays of $D_1\times{}D_2=40\times{}55$ Gabor feature vectors.
%
%
%We use Gabor features with $F=40$ entries (compare \cite{})
%
%
%We chose the 40D Gabor jets as the image patch and computed them with $3$-pixel interval, which results in a $%40\times55$ grid of Gabor jets.
% Parameter initialization
% - explaining the reason of the difference between the object size and the image segment size
%Due to the fixed interval of cutting, characters are often partly shown at boundaries of segments. For the convenience of detection, the size of segments are usually defined as twice of biggest characters.
The learning algorithm was applied to this data set assuming $C=6$ classes. The pattern mean $W$ was initialized by randomly selecting $C=6$ patches from the dataset and cutting out a segment of the pattern size at random positions. The remaining parameters were initialized in the same way as for artificial data.
To increase computational efficiency we, furthermore, assumed with $\vec{P} = (30,40)$ a pattern size smaller than the patch size but still larger than the size of any characters. Parameter optimization ($44$ EM iterations) took about $25$ minutes on a cluster with 15 GPUs (GTX 480).
%
% Show the performance
%
Fig. \ref{fig:bayes}b visualizes the inferred parameters after the
application of the learning algorithm (see Suppl. for a
visualization of the time-course of learning). As can be observed, the
algorithm has successfully represented the five character types. They
were represented by different classes using parameters for mask, mean
features and feature variances. As only five classes are needed to
represent all the characters, the algorithm has assigned a pattern averaging
other patterns and dirt to one of the classes (class $4$). In numerical experiments on
this and other documents, classes not representing
characters had either much lower values for learned mask
parameters (compare Fig.\,\ref{fig:bayes}b) or much lower values for learned mixing proportions $\pi_c$.
We exploited this observation to automatically classify character classes (see Suppl. for details).
The full learning procedure then consisted
of a repetition of the learning algorithm and a selection of one of the results
with the highest number of character classes.
%For experiments with few character types
%few repetitions proved sufficient. For instance for five character types as in Fig.\,..
%ten runs usually contained 4 to 6 representations with all characters represented.
%For more character types, e.g.\ nine as in Suppl.\ or more, all characters were often
%only represented in one or two of ten runs or in still fewer cases.
%
%The learned parameters is shown in Fig. \ref{fig:bayes}b with the visualization of the maximum of 40 feature entries. The full representation of the learned parameters and the learning course is shown in Supplement.
%
%\section{Character Detection and Identification}
%\comment{Character Classification and Identification}\\
\label{SecClassification}
\ \\[3mm]
{\bf Character detection and identification.} Based on the learned representation, characters in a given dirty document can now be detected and identified. We screen through the whole document from upper-left to lower-right patch by patch.
%Having learned a character representation, we go through the whole scanned document using overlapping patches $Y$.
Our aim is to identify a character within each patch $Y^{(n)}$ and to assign to each match a quality measure, i.e.,
a measure reporting how well each character matches the learned representation of its class. To identify the
position and type of a character in a patch we compute the MAP estimate of the approximate posterior:
{\small
\begin{equation}
\begin{split}
(c^{*}, \vec{x}^{*}) &=  \myargmax{c,\vec{x}} \{ p(c,\vec{x}\,|\,Y^{(n)},\Theta) \}\\
                &\approx \myyargmax{(c,\vec{x})\in\mathcal{K}_n}\hspace{-3mm}\{ q_n(c,\vec{x}\,;\Theta) \},
\end{split}
\label{eqn_map}
\end{equation}
}
with $q_n(c,\vec{x}\,;\Theta)$ and $\KKn$ defined as in Sec.\,\ref{SecOptimization}.
%In analogy to template matching \cite{graphmatching} we refer to the
%computation of the MAP estimate (\ref{eqn_map}) as {\em matching}, to
%$c^{\ast}$ as {\em matched class}, and to $\vec{x}^{\ast}$ as {\em matched
%position}.
%
In analogy to template matching (\cite{WiskottEtAl1997,LeCunEtAl2010} and many more) we refer to
the result of the MAP estimate (\ref{eqn_map}) as the {\em match}
for the image patch, to $\vec{x}^{\ast}$ as the {\em matched position} and to $c^{*}$ as the {\em matched class}. \comment{or {\em matched pattern}}
% Furthermore, we refer to the character pattern in the patch as {\em input pattern} in order to distinguish it from the representation of its corresponding matched pattern.
Furthermore, given the patch $Y^{(n)}$ with match $(c^{*},\vec{x}^{*})$, we define the {\em quality} of the match as follows,
%As quality measure between an input pattern and its matched pattern, we use the feature consistency measure:\vspace{-3mm}
%between input pattern features and the features of its matched pattern:
%
%As quality measure we use a distance function
%between the pattern at patch position $\vec{x}^{*}$ and the learned
%representation in pattern class $c^{*}$:
{\footnotesize
\begin{equation}
\begin{split}
Q(Y^{(n)}&,c^{*},\vec{x}^{*},\Theta) =\\ &1\,-\,\frac{\sum_{\vec{i}=(1,1)}^{(P_1,P_2)} (\alpha^{c^{*}}_{\vec{i}})^{\gamma} [\alpha^{c^{*}}_{\vec{i}} - p(m_{\vec{i}}=1\,|\,Y,c^{*},\vec{x}^{*},\Theta)]^2}{\sum_{\vec{i}'=(1,1)}^{(P_1,P_2)} (\alpha^{c^{*}}_{\vec{i}'})^{\gamma}},
\end{split}
\label{eqn_incompleteness}
\end{equation}
}
where $p(m_{\vec{i}}=1\,|\,Y,c^{*},\vec{x}^{*},\Theta)$ is the posterior distribution of the binary mask (see Eqn.\,\ref{eqn_q2}). The negative term in (\ref{eqn_incompleteness}) is a normalized distance measure between mask parameters and mask posterior probabilities. To provide some intuition, suppose
that the mask parameters are binary, i.e., they are either maximally reliable ($\alpha^{c}_{\vec{i}}=1$), or maximally unreliable ($\alpha^{c}_{\vec{i}}=0$). Then, the quality reveals the percentages of the pattern $c^*$ being matched in the patch.
If for instance a patch contains a complete and clean
instance of the pattern $c^{*}$ at position $\vec{x}^{*}$,
$p(m_{\vec{i}}=1\,|\,Y,c^{*},\vec{x}^{*},\Theta)$ is close or equal to one for all reliable features and zero otherwise. This implies that the distance measure is equal to zero and $Q(Y^{(n)},c^{*},\vec{x}^{*},\Theta)$ equal to one. For an appropriate scaling of the match quality with the degree of dirt, the
unreliable features in (\ref{eqn_incompleteness}) have been down-weighted by factors
$(\alpha^{c^{*}}_{\vec{i}})^{\gamma}$ (we use $\gamma=10$).
Given a patch $Y$ and a match $(c^{*},\vec{x}^{*})$, the measure (\ref{eqn_incompleteness}) thus assigns
a quality value $Q(Y^{(n)},c^{*},\vec{x}^{*},\Theta)\in[0,1]$ which well reflects the similarity between
an input pattern at $\vec{x}^{*}$ and its corresponding matched pattern of class $c^{*}$. Low values of $Q$ correspond
to poor matches and $Q=1$ corresponds to a perfect match (see Suppl. for details).
\section{Corrupted Document Cleaning}
\label{SecExperiments}
%
% select character objects
% determine the bounding box of characters
%
By making use of the learned character representation, character matching, and evaluation of match qualities,
we can now remove dirt from a given corrupted scanned document. First, we use the match qualities to globally find the best matching input patterns for each class among all extracted patches.
For each best input pattern we then compute a bounding box and store the corresponding pixel representation (see Fig.\,\ref{fig:bayes}c). Then, using the best representations, we can reconstruct the document (see Fig.\,\ref{fig:bayes}e).
 In order to do so, we screen through the dirty document patch by patch and for each patch compute the
match $(c^*,\vec{x}^*)$ using (\ref{eqn_map}) and the match quality using (\ref{eqn_incompleteness}).
If the matched position $\vec{x}^*$ corresponds to a pattern fully visible within the patch, and
if the match quality is above the threshold $Q_o=0.5$, we paint the best representation of
class $c^*$ at position $\vec{x}^*$ onto an initially blank reconstructed document.
Fig.\,\ref{fig:bayes}d illustrates
this procedure for a small area of the example document.
As can be observed, not all the matches are accepted for reconstruction,
because some matches correspond to patterns not entirely visible (e.g., second patch at iteration $1$) or match qualities are too low (e.g., last patch at iteration $2$).
The quality threshold prevents dirt from being reconstructed as characters. As for each patch just one
match is computed, not all characters are reconstructed at first. For a complete reconstruction we therefore
replace each successfully reconstructed character in the original
document by a blank rectangle (of the same size as the corresponding bounding box)
and apply the procedure again. Patterns that previously were not identified because
of competition with other patterns can now be found and correctly reconstructed.
We terminate the reconstruction once no more matches are accepted. In Fig.\,\ref{fig:bayes}d two iterations of the procedure are sufficient to successfully reconstruct the word ``bayes". The entire document in Fig.\,\ref{fig:bayes}e is perfectly reconstructed after three iterations. The reconstructions of examples with more character types, non-Latin characters (Klingon) and random placement of characters show similar results (see Supplement).
However, the more a document is corrupted by dirt, the less perfect we can
expect the reconstruction to be. In examples with dirt fully occluding
parts of the document, we do thus obtain many false negative errors
(see Supplement). False positive errors are, on the other hand, obtained
if, e.g., a random combination of manual line strokes coincides with
the feature arrangement of a learned pattern (see Supplement).
Although error rates for imperfect reconstructions can be decreased by
fine tuning the threshold $Q_o$, we left the parameter
unchanged at $Q_o=0.5$ for all examples to demonstrate the generality of the
approach.

Note that the task of cleaning documents such as those in
Fig.\,\ref{fig:bayes} or in the Supplement has previously not been
addressed. This is because of the difficulty posed by corruptions
consisting partly of the same features as the characters (line
strokes). Furthermore, extended line strokes severely affect any
segmentation-based processing. It is in the nature of a new
application domain that no data for comparison is available for our
results. To provide, at least, a baseline, we applied a standard OCR
approach (FineReader, \cite{FineReader}) to the documents used in our experiments.
%
%The difficulty of the task addressed is that characters and dirt
%consist of the same features: line strokes. This makes OCR
%approaches perform very poorly because they rely on
%character segmentation before classification (with classifiers
%learned with supervision). Therefore an OCR system usually
%fails for any characters significantly overlapping with line
%strokes and mistakes dirt as chars.
%
For the document of Fig.\,\ref{fig:bayes}, FineReader recognized
56.5\% of the characters correctly (essentially those that are
segmentable) and corruption by dirt causes $297$ false positives.
On the same data, our approach detects 100\% of the characters correctly
with no false positives (FP). More examples can be found in the
Supplement.  The poorest performance of FineReader in all the examples
is observed for documents with non-standard characters or unusual
character orientations. For the documents in Figs.\,11
and 15 (Suppl.  C.2 \& C.3), for instance,
FineReader results in recognition rates of 0\% (231 FP) and 0.8\% (86
FP), respectively. For comparison, our approach detects 100\% (no FP)
in Fig.\,11 and 100\% (3 FP) in
Fig.\,15.  Performance of the unsupervised
learning algorithm is high in these latter two examples because it can learn any character type
while the poor performance of FineReader is simply evidence for the
data containing characters unknown to the OCR approach. Improvements
of OCR would require additional training on labeled data. However, as briefly
discussed in the introduction, note that a comparison of OCR
to our approach on these data is not fair.  OCR is not intended for
the task addressed here.  Vice versa, our algorithm would not perform
well on typical OCR tasks.
\section{Discussion}\label{SecDiscussion}
We have studied an unsupervised approach to remove dirt from scanned
text documents. Our approach relied on the learning of character
representations using a probabilistic generative model with an explicit
position variable. Similar to other probabilistic approaches, e.g.,
image denoising, we followed the general principle of capturing the
regularities of the data, and removed unwanted data parts after
identifying them as deviations from the learned regularities. However,
in contrast to approaches for noise removal, we learned explicit
high-level representations of specific image components (characters). % (but compare \cite{}).
Having an explicit notion of feature arrangements per character allows
for a discrimination of irregular patterns vs.\ characters even though these
irregular patterns can consist of the same features (line strokes) as the characters themselves.
Methods not representing characters explicitly (e.g., \cite{Jojic2003})
are, therefore, not applicable or would, at the least, require
additional mechanisms to identify characters and to discriminate them
against irregular patterns.

%
%The technology developed for the task of document cleaning
%can, of course, also be applied more generally. By still considering
%text documents, the approach could, for instance, be used for an efficient
%`lossy' compression of the document, e.g., by simply storing the pattern
%classes, their positions and the best matching examples. Examples
%of another data domain are, for instance, microscopy images of
%human or bacterial cell growth and/or the dynamic interaction of
%such cells. Typical task for such data are counting of different cell types or
%the extraction of information about spatial cell relations.
%
%For task of document cleaning as addressed by this work, we have shown
%
%In applications to dirty documents,
%By applying the developed procedure,
%
By applying our approach we have shown in this study that even under
difficult conditions a perfect reconstruction of a document is
possible with solely the information on a single page. The result of
the cleaning procedure depended on the factors like the severity of the corruption,
the number of character instances per character type, and on the
similarity between character patterns and corrupting patterns. Very simple characters
like ``I", ``V" or ``C" are, for instance, easier to confuse with
random line strokes than more complex characters. Furthermore, the
more character types a document contains the more challenging the
discrimination between characters becomes, especially for strongly corrupted
data. This is true for learning as well as for character
identification.
%
%are used the
%
%The similarities among characters also cause problems to learning, e.g.,
%``m" and ``n". The learning algorithm will treat the data points of ``n" as the pattern ``m" being half
%occluded instead of learning it as a new pattern.
%
Regarding required data, we usually observed good result in our
experiments for more than $200$ character instances per character
type.  Performance significantly decreased for less than $100$
instances, primarily due to less appropriate learning of the character
representations. The example of Fig.\,\ref{fig:bayes}e contains about
$250$ instances per character type ($1251$ characters in total). A
page with text consisting of the full alphabet of letters, even if
constrained to just lower or upper case, would therefore not provide
sufficiently many examples for self-cleaning. A natural extension of
the addressed task for more character types would, therefore, require
several pages.  If we assume that about $200$ examples per character
type are needed and if a page contains $1000$ characters in total,
we would require about $6$ pages to learn a full Latin alphabet of
lower-case letters. For the general type-set of all letters and numbers (excluding
special characters), we would require about $13$ pages. If we, furthermore,
consider that, e.g., just $0.074\%$ of all characters in the English language
are of type `z' \cite{BekerPiper1982}, then the number of required pages
would increase to about $270$.
To execute the cleaning procedure described in this work, processing
of $270$ pages amounts to unreasonably long computation times (even
using parallel implementations).
%Furthermore, more classes would make
%the classification of characters and the discrimination of characters
%vs.\ dirt more difficult.

On the other hand, the cleaning performance can be further improved by exploiting
further regularities of text documents. The regular arrangement of
characters along a line (compare \cite{CaseyLecolinet1996}) could be used to predict
the positions of characters, and linguistic regularities (e.g., probabilistic language models)
could be used to predict character types from context.
Using probabilistic generative approaches, such prior knowledge
can be integrated into the model by constructing more sophisticated prior distribution $p(c,\vec{x}\,|\,\Theta)$.
Also on the algorithmic side improvements can certainly be made, e.g., by using a
multiple-cause structure (e.g., \cite{DayanZemel1995}) to recognize
multiple patterns in a patch simultaneously, or by using image features with scale invariance
and contrast normalization (e.g., SIFT \cite{Lowe2004}, HOG \cite{DalalTriggs2005}). Different font sizes of characters can be handled by modeling them as different patterns, adding scaling transformations to the model (dramatically increasing the computational complexity), or estimating font sizes with separate mechanisms.

By applying the probabilistic approach described in this work, we have
for the first time shown that it is in principle possible to
autonomously clean text documents which are heavily corrupted by irregular patterns.
Future developments can further improve the cleaning performance by exploiting regularities of words
and sentences, or they can extend the application domain of the approach.
%
%and integrating further
%regularities of written written text.
%
%
%exploiting higher order prior knowledge of text documents
% and by further improving  algorithms.
%
%\newpage

{
\vspace{2mm}
\noindent{\bf Acknowledgement.} This work was funded by the German Research Foundation (DFG) under grant
\mbox{LU 1196/4-1}. Early modeling work was funded by the German
Federal Ministry of Education and Research (BMBF) under grant
01GQ0840.
}

{\footnotesize
\bibliography{dirtydoc,../../bibs/jpl2011-05Plain_dirty_doc}
\bibliographystyle{ieeetr}
}

\end{document}